\documentclass[journal,twoside,web]{ieeecolor}
\usepackage{generic}
\usepackage[T1]{fontenc}
\usepackage{amsmath}
\usepackage{newtxtext}
\usepackage{newtxmath}
\DeclareFontFamilySubstitution{T1}{phv}{qhv}
\ifdefined\XeTeXversion
\DeclareFontFamily{TU}{phv}{}
\DeclareFontShape{TU}{phv}{m}{n}{<->"[texgyreheros-regular.otf]:mapping=tex-text"}{}
\DeclareFontShape{TU}{phv}{b}{n}{<->"[texgyreheros-bold.otf]:mapping=tex-text"}{}
\DeclareFontShape{TU}{phv}{m}{it}{<->"[texgyreheros-italic.otf]:mapping=tex-text"}{}
\DeclareFontShape{TU}{phv}{b}{it}{<->"[texgyreheros-bolditalic.otf]:mapping=tex-text"}{}
\DeclareFontShape{TU}{phv}{m}{sl}{<->ssub * phv/m/it}{}
\DeclareFontShape{TU}{phv}{b}{sl}{<->ssub * phv/b/it}{}
\DeclareFontShape{TU}{phv}{bx}{n}{<->ssub * phv/b/n}{}
\DeclareFontShape{TU}{phv}{bx}{it}{<->ssub * phv/b/it}{}
\DeclareFontShape{TU}{phv}{bx}{sl}{<->ssub * phv/b/it}{}
\fi

\usepackage{cite}
\usepackage{graphicx}
\usepackage{textcomp}
\usepackage{booktabs}
\usepackage{multirow}
\usepackage[table]{xcolor}
\usepackage{etoolbox}
\makeatletter
\let\NAT@parse\undefined
\makeatother
\usepackage{hyperref}
\hypersetup{colorlinks=true,linkcolor=blue,citecolor=blue,urlcolor=blue}
\definecolor{subsectioncolor}{rgb}{0,0.541,0.855}

\makeatletter
\patchcmd{\thebibliography}{plus pt}{plus 0pt}{}{}
\makeatother

\def\journalname{IEEE TRANSACTIONS ON MEDICAL IMAGING}
\newcommand{\M}{\mathcal{M}}
\newcommand{\R}{\mathbb{R}}
\newcommand{\bX}{\mathbf{X}}
\newcommand{\bR}{\mathbf{R}}
\newcommand{\bc}{\mathbf{c}}
\newcommand{\bK}{\mathbf{K}}
\newcommand{\bT}{\mathbf{T}}
\newcommand{\tS}{\tilde{S}}
\newcommand{\tr}{\tilde{r}}
\newcommand{\fage}{f_{\mathrm{age}}}
\newcommand{\fmot}{f_{\mathrm{mot}}}
\newcommand{\sedge}{s_{\mathrm{edge}}}
\newcommand{\sconf}{s_{\mathrm{conf}}}
\newcommand{\clip}{\operatorname{clip}}

\definecolor{errhead}{RGB}{251,229,214}
\definecolor{acchead}{RGB}{222,231,247}
\definecolor{oursrow}{RGB}{247,240,240}
\newcommand{\eh}[1]{\cellcolor{errhead}#1}
\newcommand{\ah}[1]{\cellcolor{acchead}#1}
\newcommand{\sbest}[1]{\underline{#1}}

\newlength{\qw}
\newlength{\qh}
\newcommand{\qlab}[1]{\rotatebox{90}{\makebox[\qh][c]{\scriptsize #1}}}
\newcommand{\qc}[1]{\includegraphics[width=\qw,height=\qh]{figures/qualitative/c3vd/#1}}
\newcommand{\qs}[1]{\includegraphics[width=\qw,height=\qh]{figures/qualitative/sm/#1}}
\newlength{\denw}
\newlength{\denh}

\begin{document}
\bstctlcite{IEEEexample:BSTcontrol}
\title{TANDEM: Temporal Anchor Densification With Reliability-Aware Memory for Lightweight Sparse Depth Completion in Endoscopic Robotic Navigation}
\author{Weikuan Qiu, Jiale Hou, Chongzhao Xu, and Jinlian Ma
\thanks{(Corresponding author: Jinlian Ma.)}
\thanks{Weikuan Qiu, Jiale Hou, Chongzhao Xu, and Jinlian Ma are with the School of Integrated Circuits, Shandong University, Jinan 250101, China (e-mail: 202300400175@mail.sdu.edu.cn; jialehou@mail.sdu.edu.cn; 202300400121@mail.sdu.edu.cn; jlma2019@sdu.edu.cn).}
\thanks{This work received no external funding.}
\thanks{This work has been submitted to the IEEE for possible publication. Copyright may be transferred without notice, after which this version may no longer be accessible.}}

\maketitle

\begin{abstract}
For endoscopic robotic surgery, accurate dense depth perception serves as a foundational capability, which enables navigation, instrument localization, and tissue interaction. However, endoscopic video lacks reliable metric depth, while metric measurements obtained via tracking, time-of-flight sensing, or instrument contact only cover a small fraction of pixels in each frame, rendering sparse depth completion an urgent and unsolved challenge. In this work, we propose Temporal Anchor Densification with Reliability-Aware Memory (TANDEM), a causal sparse depth completion framework for endoscopic video. It adopts a lightweight design and leverages the video stream: past measurements re-projected into the current view recover the metric evidence unavailable within a single sparse frame. To keep this evidence interpretable, accurate, and efficient, an explicit memory scores each re-projected measurement with analytic re-observation, age, and motion-parallax factors without overriding current measurements. A compact network then reconstructs depth via reliability-weighted gradient integration and a frequency-gated reference prior. We evaluate TANDEM on the C3VD colonoscopy and StereoMIS laparoscopy datasets. With only 5.58\,M parameters, TANDEM reduces the root mean square error by 11.7\% and 32.0\%, respectively, compared with the strongest baseline, and achieves state-of-the-art accuracy. It also remains accurate under pose errors of the magnitude reported for endoscopic pose estimation. Code will be released at https://github.com/Wk-cpu-art/TANDEM upon acceptance.

\end{abstract}


\begin{figure}[!t]
\centering
\includegraphics[width=\columnwidth]{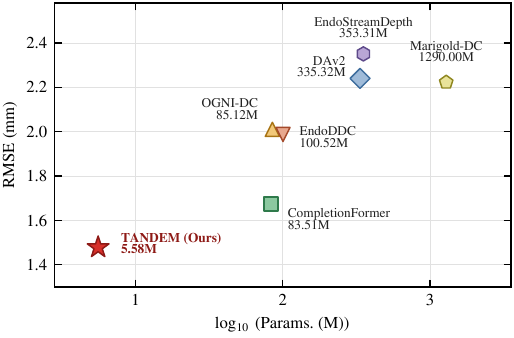}
\caption{Root mean square error versus parameter count on C3VD under very sparse input. TANDEM is both the smallest and the most accurate method.}
\label{fig:tradeoff}
\end{figure}

\section{Introduction}
\label{sec:introduction}
\IEEEPARstart{S}{urgical} robots rely on accurate three-dimensional (3-D) perception of the operating field for navigation, instrument localization, and safe tissue interaction. Endoscopic video supplies high-resolution texture, yet it does not provide stable and metrically accurate per-pixel depth. Existing routes to depth include stereo or multi-view geometry~\cite{hayoz2023stereomis}, monocular depth estimation~\cite{liu2020dense,cui2024endodac,xu2026latent,zhu2026beyond}, depth completion~\cite{11696169}, and 3-D reconstruction~\cite{yang2024deform3dgs,zhang2026sg3dgs}. In real procedures, however, each of these approaches is challenged by occlusion, specular reflection, tissue deformation, rapid viewpoint change, and a limited computational budget at the bedside. A practical alternative is therefore to anchor learned depth to the limited metric observations available during surgery and to exploit them continuously over the video stream, which motivates sparse depth completion for endoscopic navigation.

Sparse depth completion predicts a dense depth map from an image and metric measurements at scattered pixels~\cite{ma2018sparse,uhrig2017sparsity}. In endoscopy, however, metric measurements are scarce and irregular, as they originate from simultaneous localization and mapping (SLAM) or structure-from-motion landmarks on textured tissue~\cite{liu2020dense,anadon2025densification}, time-of-flight (ToF) sensing through the endoscope, or contact-based proprioception of a robotic instrument, with non-negligible acquisition cost~\cite{jordan2025probemde}. Accordingly, recent endoscopic benchmarks provide sparse depth alongside dense references~\cite{liu2025polypsense3d}, and endoscopic completion has been evaluated with tens of points per frame~\cite{11696169}. With such inputs, a single frame leaves most of the image unconstrained, so predictions fall back on appearance priors that are fragile under specular highlights and weak texture.

Fortunately, endoscopic video is a continuous stream. Measurements acquired in earlier frames remain geometrically informative while the corresponding tissue stays in view, and given camera motion from robot kinematics or tracking, they can densify the current observation at no additional sensing cost. Nevertheless, temporal evidence is not free, and its use in surgery raises three challenges. The first is \emph{temporal signal contamination}. Re-projected measurements inherit pose errors, tissue deformation, and visibility changes, and multi-frame fusion is known to blend foreground and background across depth edges under misalignment~\cite{zhu2025svdc,kim2024rayfusion}. The second is \emph{reliability interpretability}. Recurrent states and implicit feature fusion conceal which past measurement shaped a prediction, whereas a navigation system must distinguish depth that is measured now, depth that is remembered, and depth that is inferred. The third is \emph{cross-domain and scene robustness}. A temporal design that performs well on one anatomy may fail on another, even when it is trained on that anatomy. For example, a recent streaming estimator validated on colonoscopy data~\cite{li2026endostreamdepth} achieves competitive performance on a colonoscopy phantom but degrades sharply on in vivo laparoscopic sequences from our benchmark.

These observations expose a gap in current methods. On the one hand, single-frame completion networks~\cite{zhang2023completionformer,zuo2024ognidc} and generative refiners~\cite{viola2025marigolddc} discard the evidence accumulated along the endoscope trajectory. Moreover, these methods range from about 80\,M to 1.3\,B parameters (Fig.~\ref{fig:tradeoff}), which conflicts with onboard deployment. On the other hand, video depth estimators exploit time but lack metric anchoring. To close this gap, we formulate four design requirements from these three challenges: (R1) temporal measurements must be screened and weighted by explicit geometric reliability before they influence the prediction; (R2) every injected measurement must carry an inspectable provenance and confidence, and remembered measurements must never override current ones; (R3) reconstruction should be constrained by scene geometry rather than appearance alone, so that it remains dependable; and (R4) the full pipeline should be causal and compact, as a step toward onboard inference.

\begin{itemize}
\item A confidence-aware temporal memory mechanism for sparse depth completion, termed AnchorBuffer, which accumulates metric observations in a world frame and scores every remembered measurement with analytic, parameter-free reliability factors that also account for pose uncertainty, so that temporal densification remains traceable and current measurements always take precedence.
\item A geometry-guided temporal sparse-signal enhancement mechanism, which converts densified, reliability-annotated observations into an evidence-aware input representation for a compact network and couples them with reliability-weighted geometric integration and a frequency-gated reference prior.
\item A compact causal temporal depth completion framework achieving state‑of‑the‑art accuracy on the C3VD colonoscopy and StereoMIS laparoscopy benchmarks under very sparse input and remaining robust under realistic pose‑error magnitudes, with only 5.58M parameters.
\end{itemize}

\section{Related Work}

\subsection{Efficient Models for Depth Completion}
Early learning-based completion regresses dense depth directly from an image and sparse samples~\cite{ma2018sparse}, and sparsity-invariant or confidence-propagating convolutions handle irregular inputs explicitly~\cite{uhrig2017sparsity,eldesokey2020confidence}. Subsequently, spatial propagation networks refine initial estimates via learned affinities~\cite{cheng2018cspn,park2020nlspn,lin2022dyspn}. Further accuracy gains are achieved by the convolution-Transformer hybrid CompletionFormer~\cite{zhang2023completionformer}, which has a capacity of tens of millions of parameters, and by bilateral propagation starting from the earliest network stage~\cite{tang2024bpnet}. Generative models extend this trend: Marigold-DC~\cite{viola2025marigolddc} steers a diffusion-based monocular prior with sparse depth at test time, and EndoDDC~\cite{11696169} refines endoscopic completion with a diffusion model conditioned on depth-gradient features. Efficiency-oriented designs, in contrast, follow a different path. For example, FastDepth~\cite{wofk2019fastdepth} and PENet~\cite{hu2021penet} target embedded or real-time inference, and LiteSense~\cite{li2026litesense} fuses images with compact ToF histograms via patch-wise cross-attention in a lightweight encoder-decoder. In endoscopy, compact monocular estimators have been developed either through architectural design~\cite{zhang2023litemono} or by distilling geometric priors from foundation models~\cite{zhu2026beyond}. These efficient models, however, process every frame independently, which limits their accuracy when the current frame is extremely sparse.

\subsection{Geometry-Aware Sparse Depth Completion}
A complementary line injects geometric structure into completion. For instance, KBNet~\cite{wong2021kbnet} back-projects sparse points with known camera intrinsics. Wong \emph{et al.}~\cite{wong2020void} build a piecewise-planar scaffold from sparse points before refinement. OGNI-DC~\cite{zuo2024ognidc} predicts a field of log-depth gradients and integrates it via a differentiable depth integrator that treats sparse points as boundary conditions of a least-squares problem. OMNI-DC~\cite{zuo2025omnidc} extends this integrator to multiple resolutions for robustness to highly sparse and unseen patterns. Beyond geometric structure, confidence-propagating completion~\cite{eldesokey2020confidence} propagates per-pixel confidences through the network. However, these confidences describe the validity of the current input rather than their temporal provenance. Consequently, across these methods, all sparse inputs are treated as equally trustworthy observations of the current frame. None distinguishes a fresh measurement from a remembered one or adapts the geometric constraint according to the provenance of each measurement.

\subsection{Temporal Depth Completion for Endoscopic and Dynamic Scenes}
Temporal information has mostly been exploited through implicit memory. Recurrent monocular estimators propagate hidden states~\cite{patil2020dontforget}, consistent video depth estimation optimizes a scene-specific model offline~\cite{luo2020cvd}, and Video Depth Anything adds spatio-temporal heads to a foundation model~\cite{chen2025vda}. For endoscopy, EndoStreamDepth~\cite{li2026endostreamdepth} processes frames causally with multi-level Mamba temporal modules. In other domains, temporal completion methods align previous predictions or features with the current frame. For example, recurrent LiDAR completion warps past outputs with sensor poses~\cite{kaskela2024temporal}, RayFusion fuses cost volumes along rays across frames~\cite{kim2024rayfusion}, and SVDC applies frequency-selective fusion to suppress misalignment in ToF video~\cite{zhu2025svdc}. 
Overall, these approaches either keep temporal evidence implicit, which hampers attribution, or assume quasi-rigid scenes with dependable alignment. Surgical scenes, however, violate both assumptions, because tissues undergo deformation and respiratory motion, and camera motion estimates drift. Therefore, a causal mechanism that maintains explicit temporal evidence and geometrically quantifies its reliability remains missing.

\section{Methodology}

\begin{figure*}[!t]
\centering
\includegraphics[width=\textwidth]{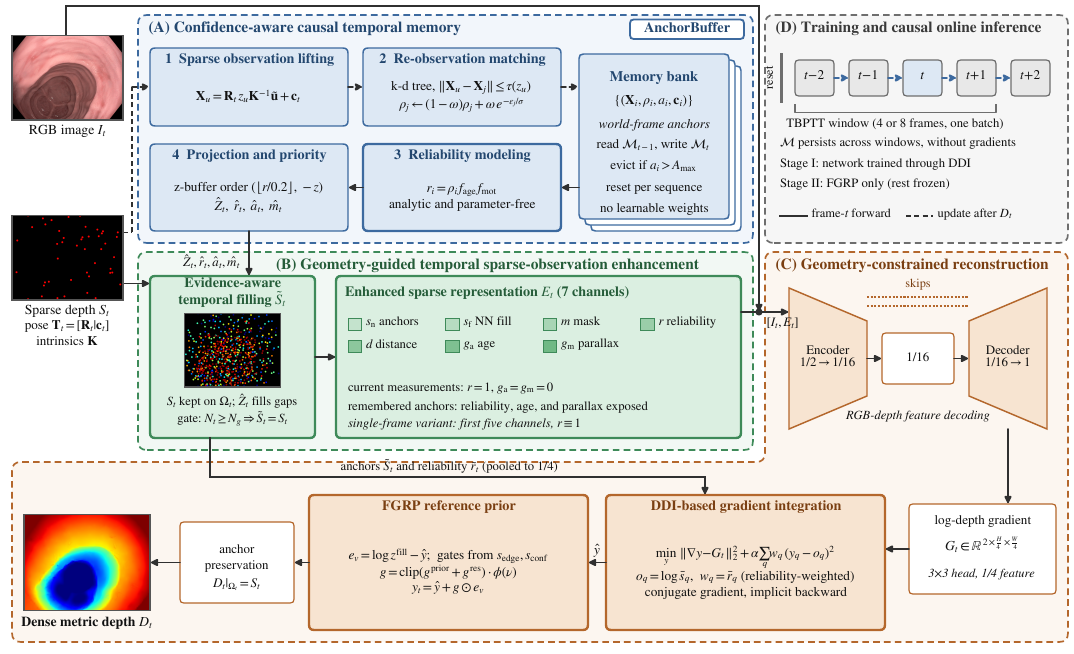}
\caption{Overview of TANDEM. (A) The memory (AnchorBuffer) is read before the network and written after the prediction. (B) Evidence-aware temporal filling completes the gaps of $S_t$ with remembered anchors under current-frame priority and encodes them into a seven-channel representation. (C) A lightweight encoder-decoder predicts a log-depth gradient field, which DDI integrates under reliability-weighted anchor constraints and FGRP refines. (D) Training and online inference follow the same causal order.}
\label{fig:framework}
\end{figure*}

\subsection{Problem Formulation and Framework Overview}
\subsubsection{Task definition}
Let $I_t\in\R^{3\times H\times W}$ denote the endoscopic image at frame $t$ and $S_t\in\R_{\geq 0}^{H\times W}$ a sparse depth map whose non-zero entries $\Omega_t=\{u\mid S_t(u)>0\}$ hold $N_t=|\Omega_t|$ metric measurements in millimeters. The camera-to-world pose $\bT_t=[\bR_t\,|\,\bc_t]\in SE(3)$ and the intrinsics $\bK$ are provided by the robot kinematics or a tracking system. Causal sparse depth completion seeks
\begin{align}
D_t&=F_\theta\!\left(I_t,S_t,\M_{t-1}\right)
\label{eq:task}\\
\M_t&=U\!\left(\M_{t-1},S_t,\bT_t\right)
\label{eq:update}
\end{align}
where $D_t\in\R_{>0}^{H\times W}$ is the dense metric depth, $\M_{t-1}$ is a memory that summarizes frames $1,\dots,t-1$, and $U$ is the memory update. Only past and current data enter \eqref{eq:task} and \eqref{eq:update}, so the prediction at frame $t$ never depends on future frames.

\subsubsection{Causal processing pipeline}
As shown in Fig.~\ref{fig:framework}, TANDEM comprises three core modules: the confidence-aware causal temporal memory (AnchorBuffer), the differentiable depth integrator (DDI), and the frequency-gated reference prior (FGRP), and it processes every frame in a fixed read-before-write order. First, the memory is read to populate the empty pixels left by $S_t$, producing the densified anchors $\tS_t$. Next, these anchors are encoded into an enhanced sparse representation that exposes the provenance of each anchor. A lightweight network then predicts log-depth gradients; the DDI integrates under reliability-weighted anchor constraints and FGRP refines into $D_t$. Finally, $S_t$ is written into the memory, which contains no learnable parameters.

\subsection{Confidence-Aware Causal Temporal Memory}
\label{sec:memory}
The memory $\M$ is a set of anchors expressed in the world coordinate frame,
\begin{equation}
\M=\left\{\left(\bX_i,\,\rho_i,\,a_i,\,\bc_i\right)\right\}_{i=1}^{|\M|}
\label{eq:memory}
\end{equation}
For the $i$th anchor, $\bX_i\in\R^3$ denotes its 3-D position, and $\rho_i\in[0,1]$ denotes its re-observation consistency. The age $a_i$ counts the frames that have elapsed since the latest observation of the anchor, and $\bc_i$ records the camera center at that observation.
Each measurement $u\in\Omega_t$ with depth $z_u=S_t(u)$ is lifted to the world frame,
\begin{equation}
\bX_u=\bR_t\,z_u\,\bK^{-1}\tilde{\mathbf{u}}+\bc_t
\label{eq:lift}
\end{equation}
where $\tilde{\mathbf{u}}$ is the homogeneous pixel coordinate.

\subsubsection{Re-observation matching}
A lifted measurement may revisit tissue that is already stored. For each $\bX_u$, the nearest stored anchor $j(u)$ is retrieved with a k-d tree, and the pair is accepted if
\begin{equation}
\left\|\bX_u-\bX_{j(u)}\right\|_2\leq\tau(z_u)=\max\left(\tau_0,\,\kappa z_u\right)
\label{eq:match}
\end{equation}
where the absolute floor $\tau_0$ and the depth-proportional term $\kappa z_u$ account for the growth of measurement uncertainty with distance. A matched anchor is confirmed by its new observation,
\begin{equation}
\rho_j\leftarrow\clip\!\left((1-\omega)\rho_j+\omega\,e^{-\epsilon_j/\sigma},\,0,\,1\right)
\label{eq:rho}
\end{equation}
where $j=j(u)$ and $\epsilon_j=\|\bX_u-\bX_j\|_2$ is the re-observation residual. The position and capture center of the anchor are then refreshed ($\bX_j\leftarrow\bX_u$, $\bc_j\leftarrow\bc_t$), and its age is reset to zero. Unmatched stored anchors age by one frame, unmatched measurements are inserted with $\rho=1$ and $a=0$, and anchors with $a_i>A_{\max}$ are evicted. If the memory still exceeds its capacity $N_{\max}$, the anchors with the lowest $\rho_i e^{-a_i/50}$ are removed. Hence, an anchor gains or loses consistency only through geometric agreement with new measurements and never through the network.

\subsubsection{Pose-aware tolerances}
Re-observation residuals also contain the error of the camera poses. When the pose source provides its uncertainty, for example, via a kinematic accuracy specification or a SLAM covariance, expressed as a rotation error $\delta_\theta$ and a translation error $\delta_t$, the expected residual of a measurement at depth $z_u$ is $\sigma_{\mathrm{pose}}(z_u)=\sqrt{(z_u\delta_\theta)^2+\delta_t^2}$, where the rotational component scales with depth. The matching threshold and the residual scale are widened accordingly,
\begin{align}
\tau'(z_u)&=\sqrt{\tau(z_u)^2+\sigma_{\mathrm{pose}}(z_u)^2}
\label{eq:poseaware}\\
\sigma'(z_u)&=\sqrt{\sigma^2+\sigma_{\mathrm{pose}}(z_u)^2}
\label{eq:poseaware2}
\end{align}
and replace $\tau$ and $\sigma$ in \eqref{eq:match} and \eqref{eq:rho}. Thus, noisy poses lower the consistency of an anchor smoothly instead of breaking its matches, whereas with exact poses $\sigma_{\mathrm{pose}}=0$ and the memory is unchanged.

\subsubsection{Reliability modeling}
When the memory is read at frame $t$, every anchor in front of the camera receives an effective reliability
\begin{equation}
r_i=\clip\!\left(\rho_i\,\fage(i)\,\fmot(i),\,0,\,1\right)
\label{eq:rel}
\end{equation}
composed of three interpretable factors. The re-observation term $\rho_i$ rewards anchors whose positions have been confirmed by independent measurements. The temporal term $\fage(i)=\exp(-a_i/T_a)$ discounts anchors that have not been observed for a long time, since the tissue may have moved. The motion-parallax term $\fmot(i)=\exp(-\beta\|\bc_t-\bc_i\|_2/z_i)$ penalizes anchors for which the difference between their capture viewpoint and the current viewpoint is large relative to their current depth $z_i$, since small pose errors and deformation can induce substantial reprojection errors. All factors are closed-form functions of stored geometric quantities, so $r_i$ is an auditable statement of why an anchor is trusted rather than a learned confidence.

\subsubsection{Projection and current-frame priority}
Each anchor is transformed from the world coordinate frame into the coordinate frame of the current camera, $\mathbf{x}_i=\bR_t^{\top}(\bX_i-\bc_t)$, and then projected to the pixel $\pi(\mathbf{x}_i)=\lfloor \bK\mathbf{x}_i/z_i\rfloor$, where $z_i$ is the depth component of $\mathbf{x}_i$. When several anchors fall on one pixel, they are ranked lexicographically by the reliability bucket $\lfloor r_i/0.2\rfloor$ and then by proximity, so that the most reliable and nearest anchor occupies the pixel, consistent with occlusion ordering. This yields maps of projected depth $\hat{Z}_t$, reliability $\hat{r}_t$, normalized age $\hat{a}_t=\min(a_i/30,1)$, and normalized parallax $\hat{m}_t=\min(\|\bc_t-\bc_i\|_2/z_i,1)$. The densified anchors give strict priority to the current frame,

\begin{equation}
\left(\tS_t(u),\,\tr_t(u)\right)=\begin{cases}\left(S_t(u),\,1\right), & u\in\Omega_t\\ \left(\hat{Z}_t(u),\,\hat{r}_t(u)\right), & \text{otherwise}\end{cases}
\label{eq:union}
\end{equation}
so remembered anchors only fill gaps and never overwrite a current measurement. Remembered anchors are injected only into sparse frames: if $N_t\geq N_g$, the memory read is bypassed and $\tS_t=S_t$, whereas the memory is still updated.
\subsection{Geometry-Guided Temporal Sparse-Observation Enhancement}
\label{sec:enhance}
\subsubsection{Enhanced sparse representation}
The densified anchors are converted into a seven-channel tensor
\begin{equation}
E_t=\left[s_{\mathrm{n}},\,s_{\mathrm{f}},\,m,\,\tr_t,\,d,\,g_{\mathrm{a}},\,g_{\mathrm{m}}\right]\in\R^{7\times H\times W}
\label{eq:enc}
\end{equation}
where $s_{\mathrm{n}}=\clip(\tS_t/Z_{\max},0,1)$ keeps the anchor values at their pixels, $m$ is the validity mask of $\tS_t$, and $\tr_t$ is the reliability map in \eqref{eq:union}. The channel $s_{\mathrm{f}}(u)=\clip(\tS_t(\nu(u))/Z_{\max},0,1)$ propagates the value of the nearest anchor $\nu(u)$, obtained with a Euclidean distance transform, so that a convolutional encoder receives a dense metric cue even when anchors are isolated. The distance channel $d(u)=\min(\|u-\nu(u)\|_2/\ell,1)$ encodes how far a pixel lies from its nearest anchor, and $g_{\mathrm{a}}=\hat{a}_t$ and $g_{\mathrm{m}}=\hat{m}_t$ expose the age and parallax of remembered anchors, both of which are zero for current measurements. The single-frame variant uses the first five channels with $\tr_t\equiv 1$.

\subsubsection{Evidence-aware temporal filling}
The representation separates three kinds of evidence: pixels in $\Omega_t$ carry measured depth with $\tr_t=1$ and zero age, pixels filled from the memory carry remembered depth with its reliability, age, and parallax, and all other pixels receive only the propagated value $s_{\mathrm{f}}$ and the distance $d$ to their supporting anchor. Since every filled pixel exposes its provenance, the subsequent modules can weight remembered evidence instead of treating it as a fresh measurement.

\subsection{Geometry-Constrained Depth Reconstruction}
\label{sec:recon}
\subsubsection{Lightweight RGB-depth feature decoding}
The encoder receives the ten-channel concatenation $[I_t,E_t]$ of the red-green-blue (RGB) image and the enhanced representation, and it follows a compact MobileNetV4-style convolutional design~\cite{qin2024mobilenetv4} with features at 1/2, 1/4, 1/8, and 1/16 resolution (32, 64, 96, and 960 channels). The first-layer weights that act on $E_t$ are initialized to zero, so training starts from an image-only network. The decoder comprises four up-sampling stages, each consisting of a $3\times3$ convolution, batch normalization, and a leaky rectified linear unit followed by a stride-2 transposed convolution; it receives the encoder features at 1/2 and 1/4, projected by $1\times1$ convolutions to 128 and 256 channels, and the encoder features at 1/8 and 1/16 as skip connections. Two outputs are produced: an auxiliary base depth $D^{\mathrm{b}}_t$ from a softplus head at full resolution and a log-depth gradient field $G_t=(G^x_t,G^y_t)\in\R^{2\times\frac{H}{4}\times\frac{W}{4}}$ from a $3\times3$ convolution on the 1/4-resolution decoder feature. The network thus predicts local geometry, while metric scale is supplied by the anchors.

\subsubsection{DDI-based gradient integration}
Following OGNI-DC~\cite{zuo2024ognidc}, DDI converts $G_t$ into log-depth $y\in\R^{\frac{H}{4}\times\frac{W}{4}}$ under anchor constraints. Anchors are pooled by masked averaging over $4\times4$ cells: for a cell $q$ with coverage $\kappa_q>0$, $\bar{s}_q$ is the mean depth of its anchors in $\tS_t$, $o_q=\log\bar{s}_q$, and $\bar{r}_q$ is the mean anchor reliability. The integrated solution minimizes
\begin{equation}
\begin{aligned}
\mathcal{E}(y)=&\sum_{d\in\{x,y\}}\ \sum_{(p,q)\in\mathcal{N}_d}\left(y_q-y_p-G^d_t(q)\right)^2\\
&+\alpha\sum_{q:\,\kappa_q>0}w_q\left(y_q-o_q\right)^2 
\end{aligned}
\label{eq:energy}
\end{equation}
where $w_q=\bar{r}_q$, $\mathcal{N}_x$ and $\mathcal{N}_y$ contain horizontally and vertically adjacent cell pairs $(p,q)$, and $\alpha$ balances anchor fidelity against gradient consistency. Cells covered only by current measurements ($\tr_t=1$) receive the full weight, whereas cells supported by weakly reliable remembered anchors are smoothly attenuated rather than discarded. With the finite-difference operator $\mathbf{A}$, the stacked gradients $\mathbf{g}_t$, and $\mathbf{W}=\operatorname{diag}(w_q\,\mathbf{1}[\kappa_q>0])$, the minimizer satisfies the sparse symmetric positive-definite system
\begin{equation}
\left(\mathbf{A}^{\top}\mathbf{A}+\alpha\mathbf{W}\right)y^{*}=\mathbf{A}^{\top}\mathbf{g}_t+\alpha\mathbf{W}\mathbf{o}
\label{eq:normal}
\end{equation}
which is solved by the conjugate gradient (CG) method. Gradients follow from implicit differentiation~\cite{amos2017optnet} with a second CG solve. The solution is bilinearly up-sampled, $\hat{y}_t=\operatorname{Up}(y^{*})$. Since every predicted gradient is reconciled with all anchors through \eqref{eq:normal}, errors in $G_t$ cannot shift the metric scale, and remembered anchors act as boundary conditions where the current frame provides none.

\subsubsection{FGRP reference prior}
Integration at 1/4 resolution smooths depth discontinuities, whereas the anchors carry sharp local evidence. To restore such discontinuities, FGRP decides where the final log-depth should lean toward the nearest anchor. Given the nearest-anchor depth $z^{\mathrm{fill}}=Z_{\max}s_{\mathrm{f}}$, it defines the evidence residual
\begin{equation}
e_v=\clip\!\left(\log z^{\mathrm{fill}}-\hat{y}_t,\,-1,\,1\right)
\label{eq:ev}
\end{equation}
and two analytic context fields: $\sedge$, the standard deviation of $\log z^{\mathrm{fill}}$ in a $5\times5$ window divided by 0.35, and $\sconf$, the reliability of the nearest anchor obtained by propagating $\tr_t$. With the frequency response $\nu=\sedge/(\sedge+1)$, the explicit prior gate is
\begin{equation}
g^{\mathrm{prior}}=\operatorname{sg}\!\left[\clip\!\left(\tanh\!\left(k\,\nu\left(\sconf-\tau_c\right)\right),0,1\right)\right]
\label{eq:gprior}
\end{equation}
where $\operatorname{sg}[\cdot]$ stops gradients. It vanishes in smooth regions, opens near anchor-supported depth edges, and remains closed for unreliable anchors. A residual network $\Phi$ with two $3\times3$ convolution layers reads $[E_t,\mathbf{1}_{\Omega_t},\hat{y}_t,\log z^{\mathrm{fill}},e_v,\sedge,\sconf,I_t]$ and predicts a gate correction
\begin{equation}
g^{\mathrm{res}}=\tanh\!\left(\psi_g(\Phi(\cdot))+b\,\sedge\sconf\right)
\label{eq:gres}
\end{equation}
with a zero-initialized $3\times3$ convolution head $\psi_g$ and a scalar $b$ initialized to zero. The final prediction is
\begin{align}
g&=\clip\!\left(g^{\mathrm{prior}}+g^{\mathrm{res}},-1,1\right)\cdot\clip\!\left(\nu/\nu_0,0,1\right)
\label{eq:gate}\\
y_t&=\hat{y}_t+g\odot e_v,\qquad D_t=\exp(y_t)
\label{eq:final}
\end{align}
where the last factor in \eqref{eq:gate} suppresses gate responses in flat regions with $\nu<\nu_0$. Finally, $D_t(u)=S_t(u)$ for $u\in\Omega_t$, so the output reproduces every current measurement. Since $g^{\mathrm{prior}}$ is fixed and the learnable component only adds a bounded correction, whether a pixel adheres to an anchor is governed primarily by the measured edge strength and the anchor reliability. FGRP constrains the prediction to shift only within the bounded residual $e_v$ toward the nearest anchor.

\subsection{Training Objective and Causal Online Inference}
\subsubsection{Training objectives}
The dense loss combines a scale-invariant logarithmic term~\cite{eigen2014depth} with a Sobel gradient term,
\begin{equation}
\begin{aligned}
\mathcal{L}_{\mathrm{dense}}(D)=\ &5\sqrt{\operatorname{Var}(\boldsymbol{\ell})+0.15\,\bar{\ell}^{\,2}}\\
&+\tfrac{1}{2}\left\langle\left|\nabla_{\!S}D-\nabla_{\!S}D^{*}\right|\right\rangle
\end{aligned}
\label{eq:dense}
\end{equation}
where $\boldsymbol{\ell}=\log D-\log D^{*}$ over valid ground-truth pixels, $\bar{\ell}$ is its mean, $\nabla_{\!S}$ denotes the Sobel operators, and $\langle\cdot\rangle$ is a spatial mean. Current and remembered anchors are supervised separately,
\begin{equation}
\begin{aligned}
\mathcal{L}_{\mathrm{anc}}(D)&=\lambda_a\left\langle|D-S_t|\right\rangle_{\Omega_t}\\
\mathcal{L}_{\mathrm{mem}}(D)&=\lambda_m\left\langle|D-\tS_t|\right\rangle_{\Omega^{\mathrm{mem}}_t}
\end{aligned}
\label{eq:anc}
\end{equation}
where $\Omega^{\mathrm{mem}}_t$ contains remembered anchors with $\tr_t>0.5$, so the network is pulled only gently toward reliable memory. During training, the anchor constraints of DDI in \eqref{eq:energy} are likewise restricted to $\Omega_t\cup\Omega^{\mathrm{mem}}_t$, whereas at inference all remembered anchors enter them with their reliability weights. This design avoids supervision being skewed by highly uncertain temporal anchors during training. In inference, all remembered anchors are retained and weighted by their reliability, which allows weak temporal evidence to contribute gradually, rather than receiving the same trust as newly‑observed current‑frame measurements. The gradient head is aligned with the ground-truth log-depth $\log D^{*}_{\downarrow4}$ on the integration grid through first-order differences,
\begin{equation}
\mathcal{L}_{G}=\sum_{d\in\{x,y\}}\left\langle\left|\Delta_d G^d_t-\Delta_d\log D^{*}_{\downarrow4}\right|\right\rangle
\label{eq:lg}
\end{equation}
over fully valid cells. Training has two stages. Stage~I optimizes all parameters except those of FGRP with
\begin{equation}
\begin{aligned}
\mathcal{L}_{\mathrm{I}}=\ &\mathcal{L}_{\mathrm{dense}}(D_t)+\mathcal{L}_{\mathrm{mem}}(D_t)+\lambda_G\mathcal{L}_{G}\\
&+\lambda_b\left[\mathcal{L}_{\mathrm{dense}}(D^{\mathrm{b}}_t)+\mathcal{L}_{\mathrm{anc}}(D^{\mathrm{b}}_t)\right]
\end{aligned}
\label{eq:stage1}
\end{equation}
where the anchor term acts on the base output because $D_t$ reproduces the current measurements by construction. Stage~II freezes these parameters, including batch-normalization statistics, and trains FGRP with $\mathcal{L}_{\mathrm{dense}}(D_t)+\mathcal{L}_{\mathrm{mem}}(D_t)+\lambda_{\mathrm{gate}}\mathcal{L}_{\mathrm{gate}}$, where
\begin{align}
\mathcal{L}_{\mathrm{gate}}&=\frac{\sum\varpi\left(\tilde{g}-g^{*}\right)^2}{\sum\varpi}
\label{eq:lgate}\\
g^{*}&=\clip\!\left(\frac{\log D^{*}-\hat{y}_t}{e_v},-1,1\right)
\label{eq:gstar}
\end{align}
Here, $\tilde{g}$ is the gate before flat-region suppression, $\varpi=\nu\,\sconf$ concentrates the term on anchor-supported edges, and pixels with $|e_v|\leq0.05$ are excluded. The target $g^{*}$ is the fraction of $e_v$ that moves the prediction toward the ground truth, so the gate learns to separate anchors on the same side of an edge from anchors that bleed across it.

\subsubsection{Online inference and memory update}
Training mirrors deployment. Sequences are visited in random order, whereas frames within a sequence are streamed in temporal order and grouped into truncated back-propagation through time (TBPTT) windows of four or eight frames, each forwarded as one batch. The memory persists across the windows of a sequence, is reset when a new sequence begins, and receives detached measurements only. At inference, the same read-before-write order is executed frame by frame.

\subsubsection{Computational complexity}
TANDEM has 5.58\,M parameters: 1.26\,M in the encoder, 4.29\,M in the decoder, 21\,K in the skip projections, 2.3\,K in the gradient head, and 4.9\,K in FGRP. At $320\times256$, the network and FGRP require 12.8 and 0.78 giga floating-point operations (GFLOPs), respectively, whereas DDI and the memory have no parameters. The memory is also bounded: every anchor is refreshed or aged at each frame and evicted after $A_{\max}$ frames, so a steady input of $N_t$ measurements keeps at most $(A_{\max}+1)N_t$ anchors, e.g., 1,550 in the sparsest setting of Section~\ref{sec:experiments} with $A_{\max}=30$. Hence, the memory scales with the number of measurements rather than with the image resolution.

\begin{table*}[!t]
\centering
\caption{Comparison Under Very Sparse Input (50 Points per Frame; RMSE and MAE in mm). Best in Bold, Second Best Underlined}
\label{tab:main}
\setlength{\tabcolsep}{4.7pt}
\renewcommand{\arraystretch}{1.05}
\footnotesize
\begin{tabular}{l c | c c c c | c c c c}
\toprule
\multirow{2}{*}{Method} & \multirow{2}{*}{Params (M) $\downarrow$} & \multicolumn{4}{c|}{C3VD~\cite{bobrow2023c3vd}} & \multicolumn{4}{c}{StereoMIS~\cite{hayoz2023stereomis}}\\
 & & \eh{RMSE $\downarrow$} & \eh{MAE $\downarrow$} & \eh{REL $\downarrow$} & \ah{$\delta$ $\uparrow$} & \eh{RMSE $\downarrow$} & \eh{MAE $\downarrow$} & \eh{REL $\downarrow$} & \ah{$\delta$ $\uparrow$}\\
\midrule
\multicolumn{10}{l}{\emph{Monocular depth estimation}}\\
DAv2~\cite{yang2024dav2} & 335.32 & 2.240 & 1.663 & 0.0879 & 0.9280 & 9.619 & 6.856 & 0.2111 & 0.6185\\
EndoStreamDepth~\cite{li2026endostreamdepth} & 353.31 & 2.351 & 1.458 & 0.0815 & 0.9078 & 11.163 & 7.107 & 0.2158 & 0.5957\\
\midrule
\multicolumn{10}{l}{\emph{Sparse depth completion}}\\
CompletionFormer~\cite{zhang2023completionformer} & \sbest{83.51} & \sbest{1.674} & \sbest{0.841} & \sbest{0.0405} & \sbest{0.9841} & 4.877 & 3.013 & 0.1041 & \sbest{0.8917}\\
OGNI-DC~\cite{zuo2024ognidc} & 85.12 & 2.010 & 1.098 & 0.0508 & 0.9774 & \sbest{4.591} & 3.014 & 0.1058 & 0.8884\\
Marigold-DC~\cite{viola2025marigolddc} & 1290.00 & 2.223 & 0.999 & 0.0464 & 0.9750 & 4.877 & 3.333 & 0.1177 & 0.8591\\
EndoDDC~\cite{11696169} & 100.52 & 1.987 & 0.941 & 0.0422 & 0.9802 & 4.648 & \sbest{2.991} & \sbest{0.1035} & 0.8900\\
\midrule
\rowcolor{oursrow}\textbf{TANDEM (Ours)} & \textbf{5.58} & \textbf{1.478} & \textbf{0.823} & \textbf{0.0373} & \textbf{0.9890} & \textbf{3.121} & \textbf{2.205} & \textbf{0.0787} & \textbf{0.9502}\\
\bottomrule
\end{tabular}
\end{table*}

\begin{figure}[!t]
\centering
\setlength{\tabcolsep}{0.3pt}
\renewcommand{\arraystretch}{0}
\begin{tabular}{@{}c@{\hspace{1.5pt}}cc@{\hspace{2.5pt}}cc@{\hspace{1.5pt}}l@{}}
 & \multicolumn{2}{c}{\rule{0pt}{8pt}\footnotesize C3VD} & \multicolumn{2}{c}{\footnotesize StereoMIS} & \\[3pt]
\qlab{Input} & \qc{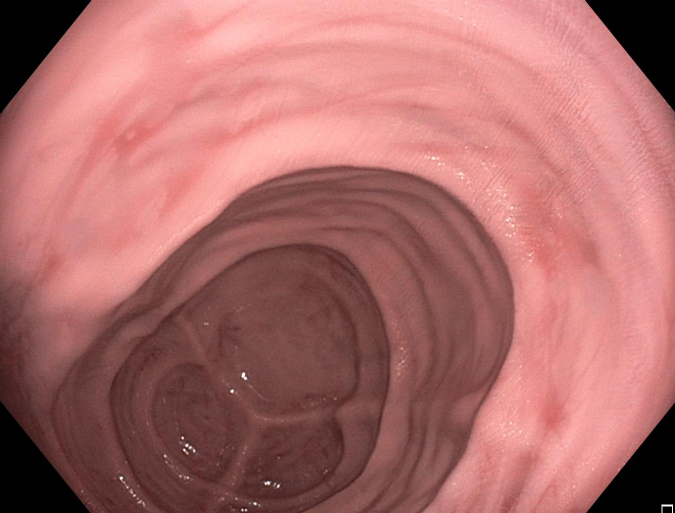} & \qc{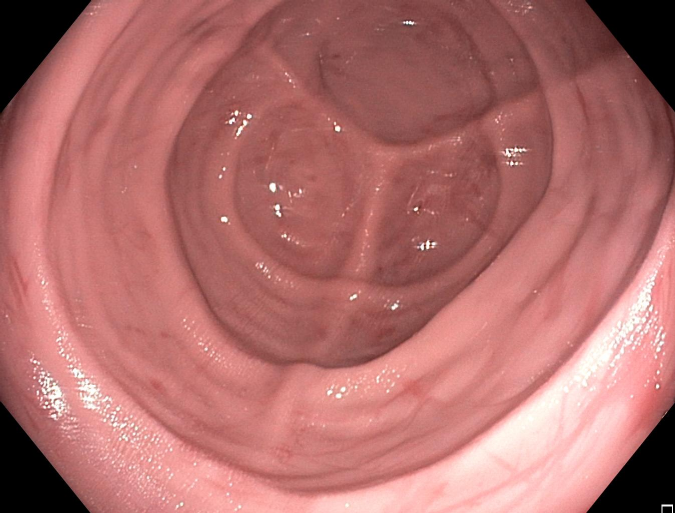} & \qs{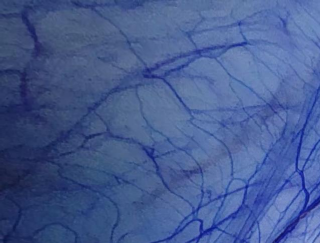} & \qs{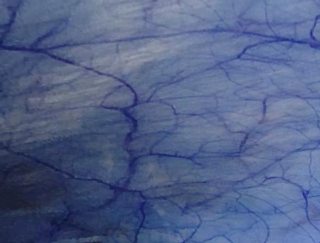} & \\[1pt]
\qlab{GT} & \qc{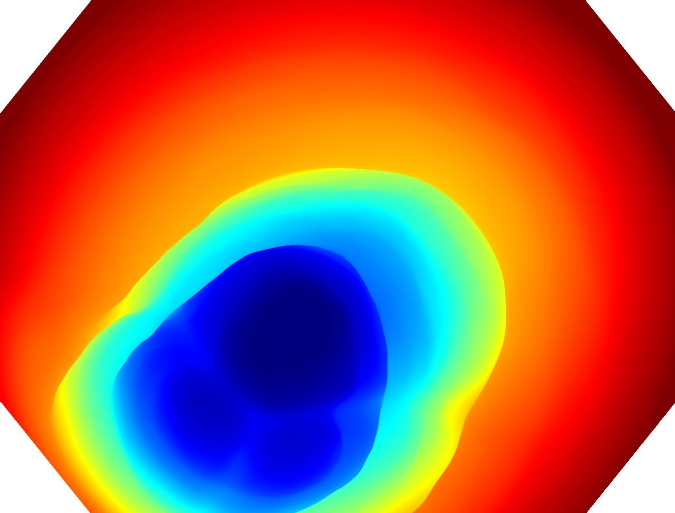} & \qc{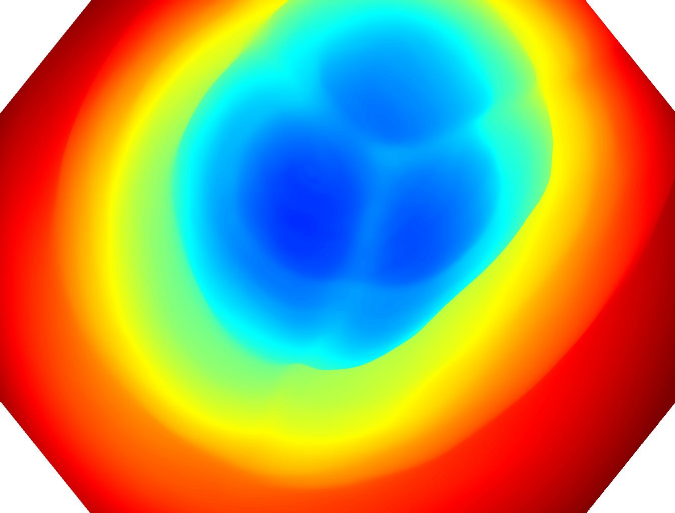} & \qs{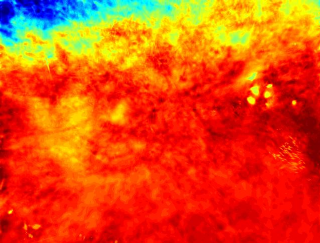} & \qs{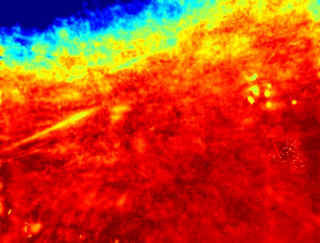} & \\[1pt]
\qlab{Marigold-DC} & \qc{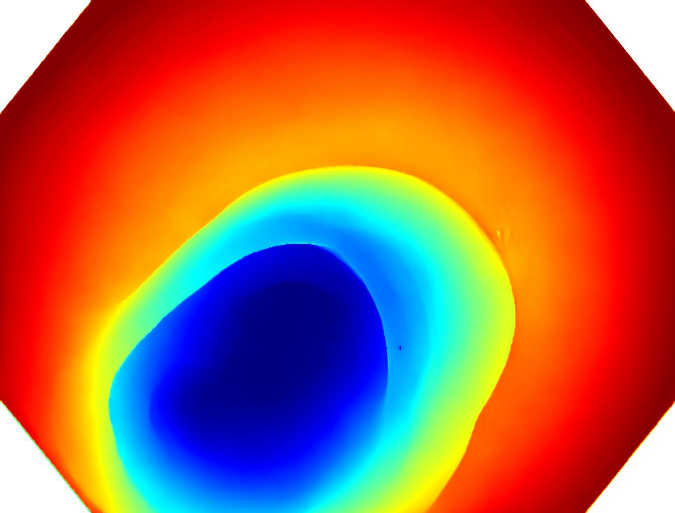} & \qc{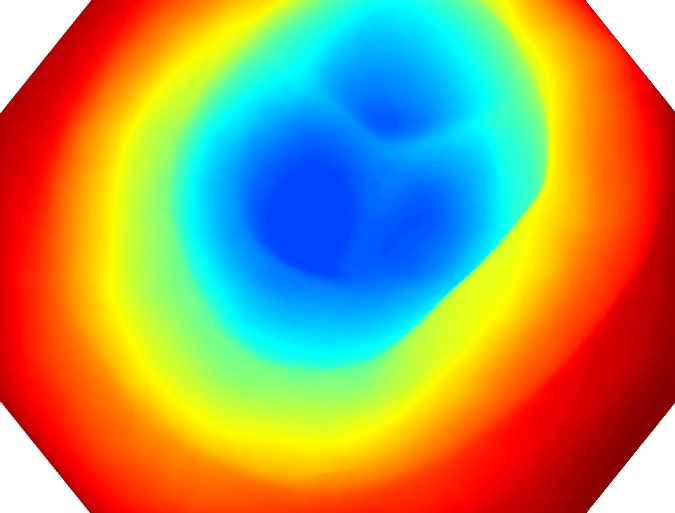} & \qs{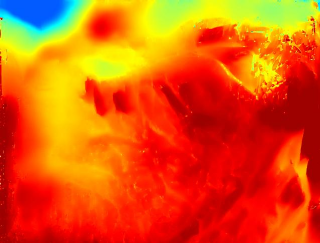} & \qs{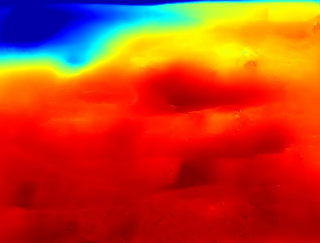} & \\[1pt]
\qlab{Error} & \qc{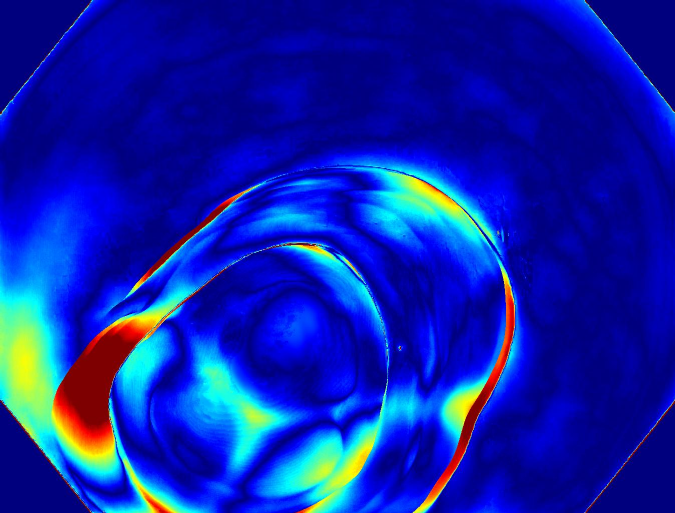} & \qc{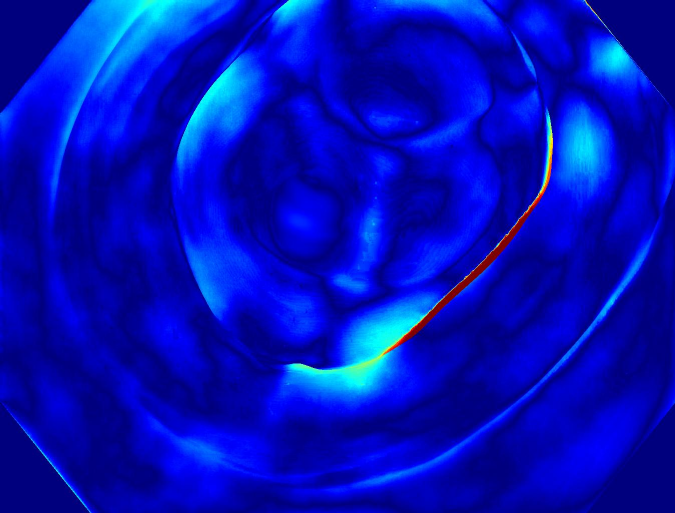} & \qs{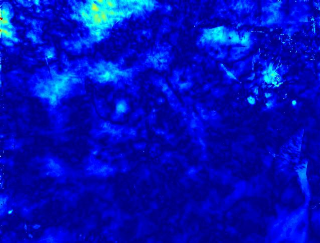} & \qs{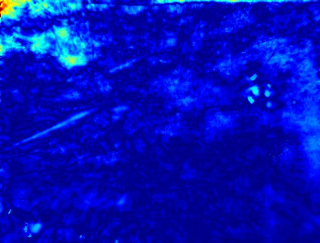} & \\[1pt]
\qlab{EndoDDC} & \qc{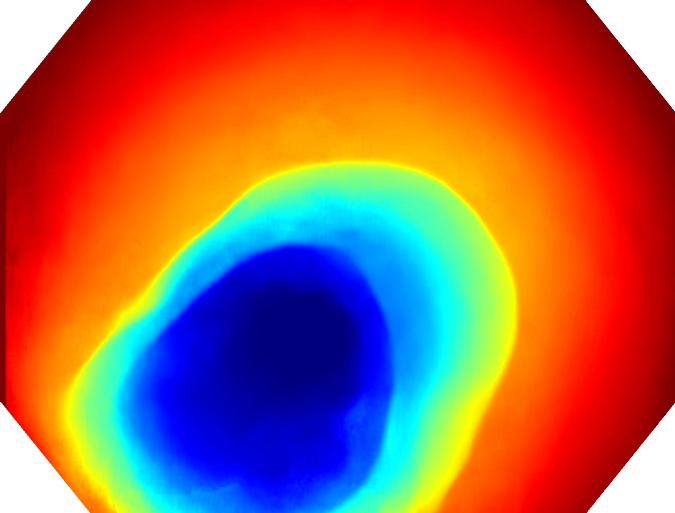} & \qc{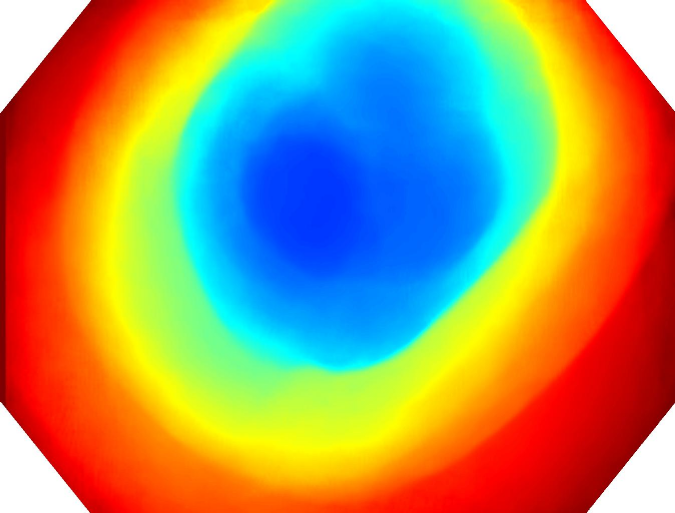} & \qs{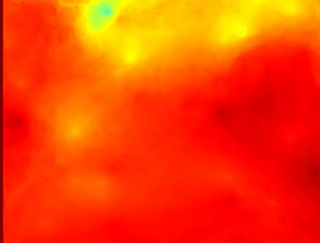} & \qs{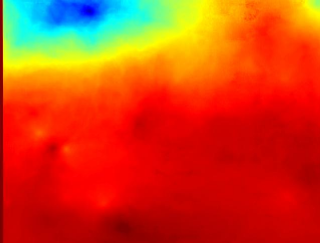} & \\[1pt]
\qlab{Error} & \qc{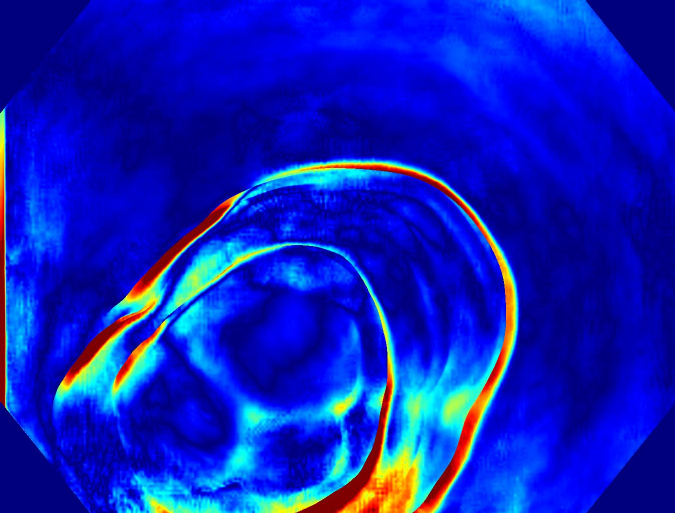} & \qc{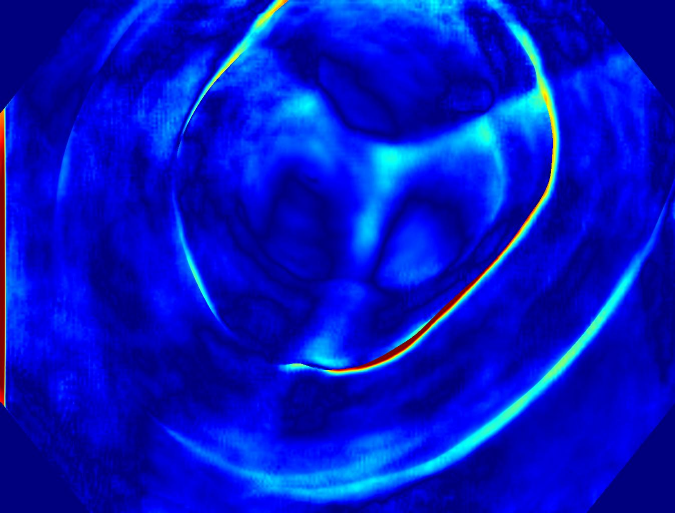} & \qs{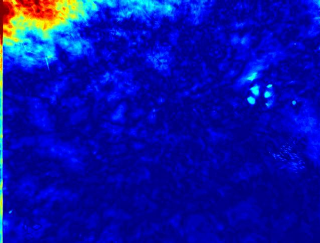} & \qs{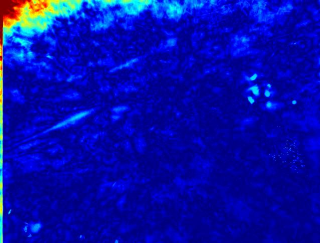} & \\[1pt]
\qlab{\shortstack{EndoStream\\Depth}} & \qc{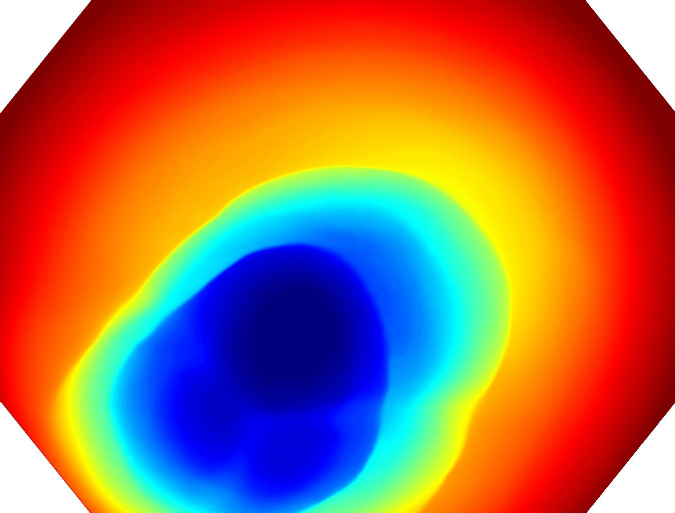} & \qc{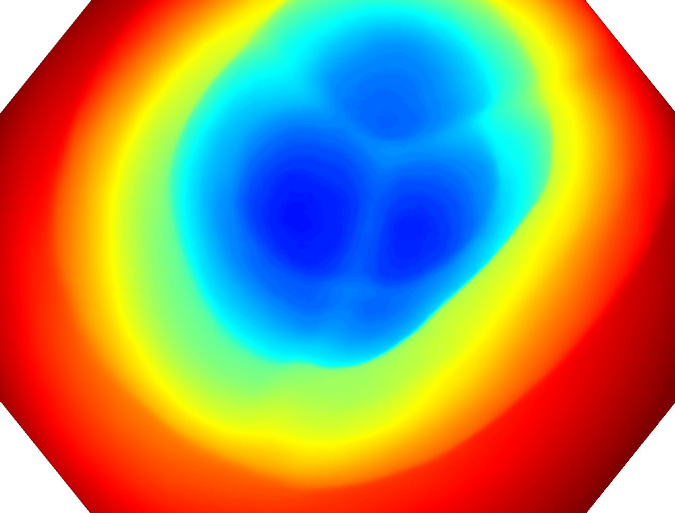} & \qs{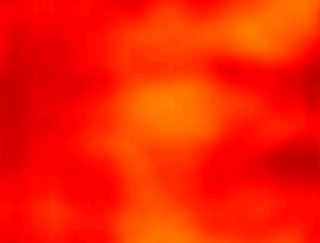} & \qs{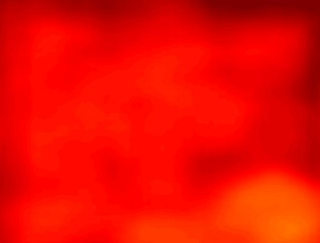} & \\[1pt]
\qlab{Error} & \qc{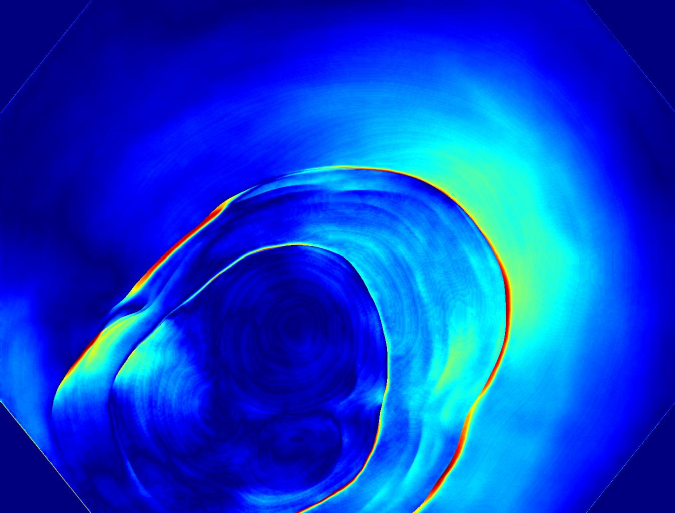} & \qc{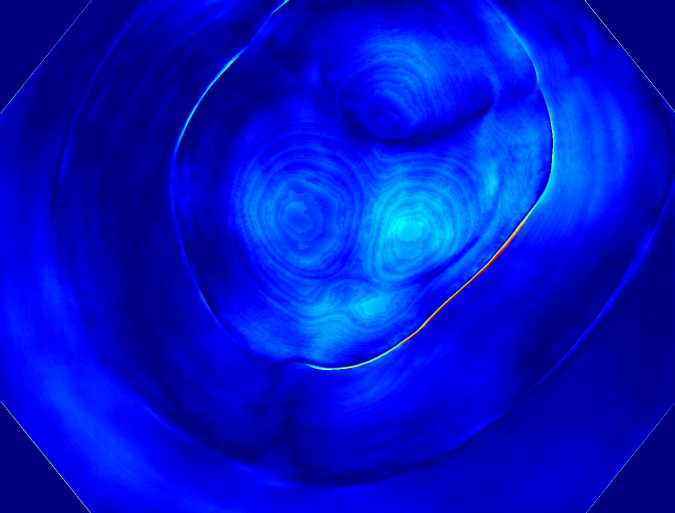} & \qs{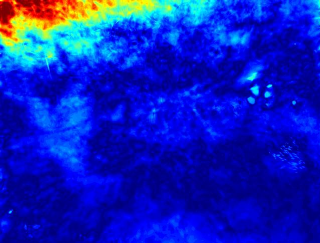} & \qs{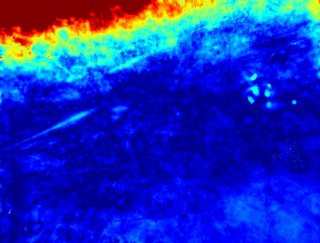} & \\[1pt]
\qlab{\textbf{TANDEM}} & \qc{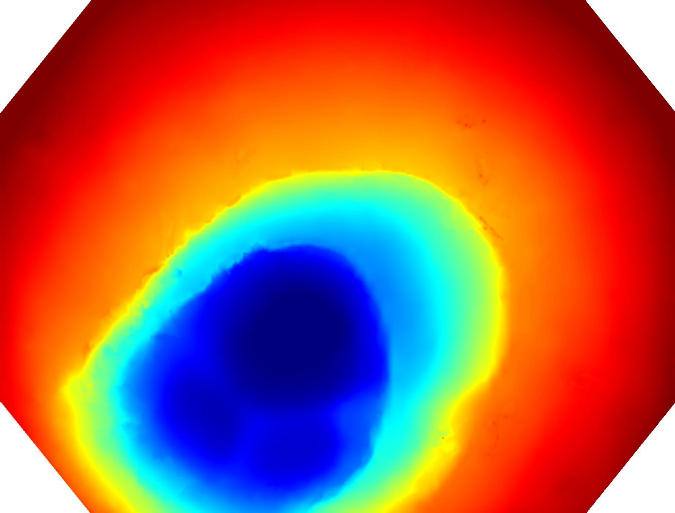} & \qc{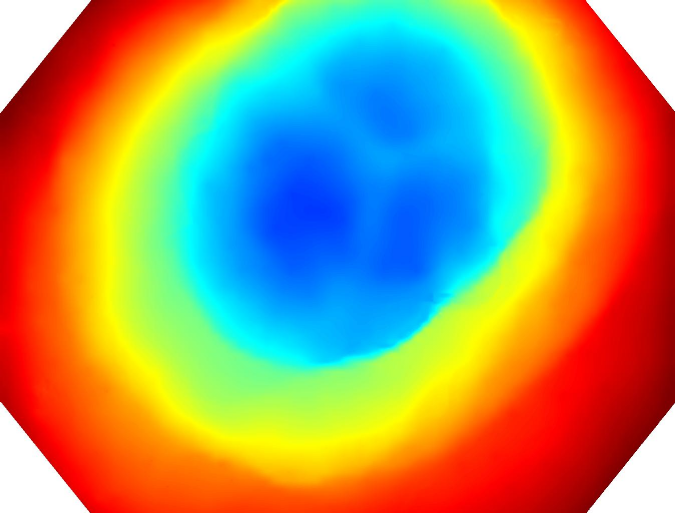} & \qs{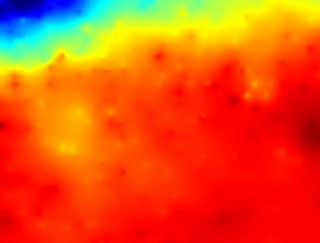} & \qs{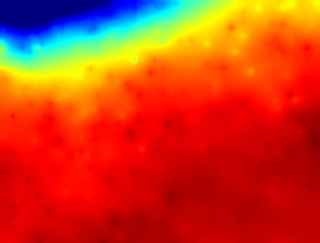} & \includegraphics[height=\qh]{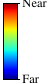}\\[1pt]
\qlab{Error} & \qc{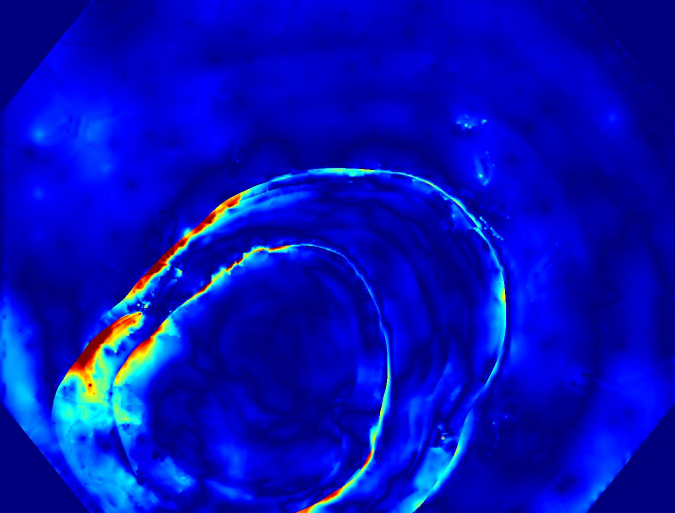} & \qc{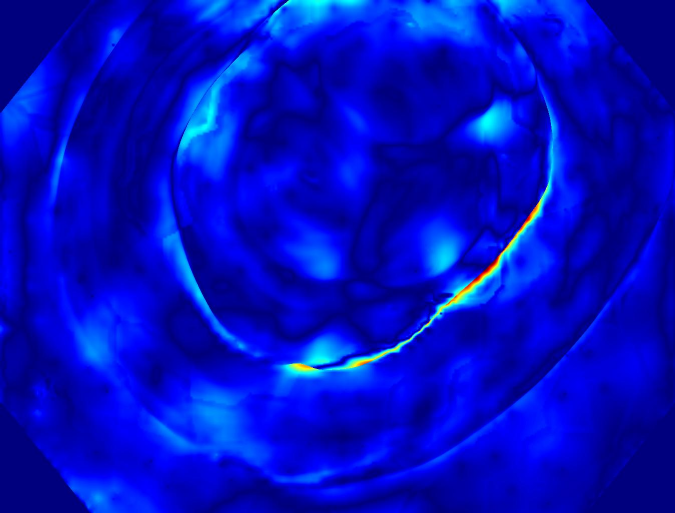} & \qs{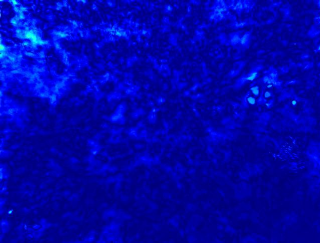} & \qs{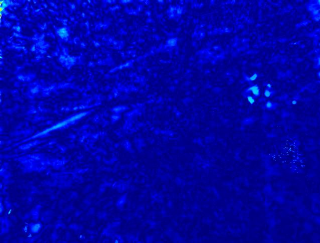} & \includegraphics[height=\qh]{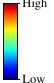}\\
\end{tabular}
\caption{Qualitative comparison under very sparse input on two C3VD (left) and two StereoMIS frames (right). Within each dataset, depth maps (red: near, blue: far) and absolute error maps (blue: low, red: high) share one color scale.}
\label{fig:qual}
\end{figure}

\section{Experiments}
\label{sec:experiments}

\subsection{Experimental Settings}
\subsubsection{Datasets}
C3VD~\cite{bobrow2023c3vd} contains colonoscopy videos of silicone colon phantoms with registered ground-truth depth and camera poses. We followed the split of LightDepth~\cite{rodriguez2023lightdepth}, with 8,690 training, 148 validation, and 1,177 test frames. We resized frames from $1350\times1080$ to $320\times256$, and clamped depth values to 100\,mm. StereoMIS~\cite{hayoz2023stereomis} contains in vivo porcine sequences recorded with a da~Vinci Xi system. Following EndoDDC~\cite{11696169}, reference depth is rendered from a Deform3DGS reconstruction~\cite{yang2024deform3dgs} on the rectified left view, and camera poses and per-scene intrinsics are expressed in the same rectified frame. The training, validation, and test sets contain disjoint scenes, and close-range training frames are oversampled three times. For both datasets, sparse inputs are drawn uniformly at random from valid ground-truth pixels. 

\subsubsection{Implementation details}
TANDEM adopts a two-stage training strategy. Stage~I is trained for 36 epochs with AdamW~\cite{loshchilov2019adamw}, an initial learning rate of $5\times10^{-4}$, a weight decay of $10^{-2}$, three warm-up epochs followed by cosine decay to $10^{-5}$, and gradient clipping at 1.0. Stage~II trains FGRP for 24 epochs with a learning rate of $10^{-3}$ and one warm-up epoch. Following a mixed-density strategy, the number of points of every training frame is drawn from $\{50, 500, 5000\}$; anchors are randomly dropped at a rate of 10\%--30\% and perturbed by multiplicative noise of at most 1\%, and photometric augmentation is applied without geometric transforms to preserve pose consistency. The checkpoint with the lowest validation root mean square error (RMSE) is retained. We set $\tau_0=2$\,mm, $\kappa=0.005$, $\sigma=3$\,mm, $\omega=0.5$, $A_{\max}=30$, $T_a=8$, $\beta=1.5$, $N_{\max}=2\times10^4$, $N_g=1000$, $\ell=32$ pixels, $\lambda_a=0.5$, $\lambda_m=0.05$, $\lambda_b=\lambda_G=\lambda_{\mathrm{gate}}=1$, $\alpha=5$, $k=4$, $\tau_c=0.5$, and $\nu_0=0.05$, and the CG solver stops after 500 iterations or at a relative tolerance of $10^{-5}$. The pose-aware tolerances \eqref{eq:poseaware} and \eqref{eq:poseaware2} use $\delta_\theta=\delta_t=0$ unless pose noise is studied. The two datasets share all of these values and differ only in the depth range, $Z_{\max}=100$\,mm for C3VD and 250\,mm for StereoMIS. At inference, $G_t$ is clipped to $[-2,2]$ before integration as a numerical safeguard, which leaves all C3VD results unchanged and takes effect only on StereoMIS with 5,000 or more input points, beyond the training range. 

\begin{table}[!t]
\centering
\caption{Temporal Densification Across Input Densities. \#Anchors Is the Mean Number of Anchors After the Priority Union, and Median $r$ Is the Median Reliability of Remembered Anchors}
\label{tab:density}
\setlength{\tabcolsep}{4.6pt}
\renewcommand{\arraystretch}{1.05}
\footnotesize
\begin{tabular}{c c c c c c}
\toprule
Input points & \#Anchors & Ratio & Median $r$ & \eh{RMSE $\downarrow$} & \eh{REL $\downarrow$}\\
\midrule
\multicolumn{6}{l}{\emph{C3VD~\cite{bobrow2023c3vd}}}\\
50 & 456 & 9.50$\times$ & 0.328 & 1.478 & 0.0373\\
500 & 1,482 & 3.08$\times$ & 0.630 & 0.879 & 0.0166\\
5,000 & 4,805 & 1.00$\times$ & -- & 0.776 & 0.0098\\
50,000 & 48,047 & 1.00$\times$ & -- & 0.600 & 0.0070\\
\midrule
\multicolumn{6}{l}{\emph{StereoMIS~\cite{hayoz2023stereomis}}}\\
50 & 661 & 13.90$\times$ & 0.260 & 3.121 & 0.0787\\
500 & 2,328 & 4.91$\times$ & 0.507 & 2.538 & 0.0621\\
5,000 & 4,746 & 1.00$\times$ & -- & 1.657 & 0.0412\\
50,000 & 47,461 & 1.00$\times$ & -- & 0.910 & 0.0216\\
\bottomrule
\end{tabular}
\end{table}

\subsection{Comparisons With Related Methods and Evaluation Metrics}
We selected two types of methods for comparison, including sparse depth completion (CompletionFormer~\cite{zhang2023completionformer}, OGNI-DC~\cite{zuo2024ognidc}, EndoDDC~\cite{11696169}, and Marigold-DC~\cite{viola2025marigolddc}), monocular depth estimation (Depth Anything V2 (DAv2)~\cite{yang2024dav2} and EndoStreamDepth~\cite{li2026endostreamdepth}). We reported the RMSE (mm), mean absolute error (MAE, mm), mean absolute relative error (REL), and threshold accuracy $\delta$, the fraction of pixels with $\max(D/D^{*},D^{*}/D)<1.25$. Metrics are computed per frame over pixels with $D^{*}>10^{-4}$\,mm and averaged over the test set, and all methods are evaluated within the same $228\times304$ evaluation region defined by the code released with EndoDDC. The primary setting provides 50 points per frame.
\subsubsection{Quantitative results}
Table~\ref{tab:main} summarizes the results under very sparse input. TANDEM achieves the best value for all eight accuracy metrics while having the fewest parameters. On C3VD, it reduces the RMSE of the strongest competitor, CompletionFormer, from 1.674\,mm to 1.478\,mm (11.7\%), together with 2.1\% lower MAE and 7.9\% lower REL. On StereoMIS, the RMSE decreases from 4.591\,mm (OGNI-DC) to 3.121\,mm (32.0\%), and MAE and REL fall by 26.3\% and 24.0\% relative to the best competing values.

The comparison reflects the challenges that motivate this work. First, the single-frame completion networks cannot draw on temporal evidence. Although CompletionFormer, OGNI-DC, and EndoDDC are trained with the same mixed-density input distribution as TANDEM, their RMSE under the same sparse input is 1.13--1.36 times that of TANDEM on C3VD and 1.47--1.56 times on StereoMIS. In addition, the zero-shot Marigold-DC, despite its 1290.00\,M parameters, remains less accurate than the trained completion networks on C3VD. Second, the monocular estimators reveal the robustness challenge. EndoStreamDepth reaches an RMSE of 2.351\,mm on C3VD, yet it yields the highest StereoMIS RMSE of 11.163\,mm, and its $\delta$ drops from 0.9078 to 0.5957; DAv2 follows the same pattern, with $\delta$ decreasing from 0.9280 to 0.6185. Although trained on each dataset separately, both estimators lack metric anchors and therefore deteriorate in the in vivo laparoscopic scenes with a wider depth range, tissue deformation, and specular reflection. In contrast, TANDEM maintains a $\delta$ of 0.9890 on C3VD and 0.9502 on StereoMIS, while having 15.0 times fewer parameters than the smallest competitor.

\subsubsection{Qualitative results}
Fig.~\ref{fig:qual} compares predictions and absolute error maps on two C3VD and two StereoMIS frames, and each competitor fails in a characteristic way. Marigold-DC recovers plausible colon geometry but accumulates error along haustral folds. On StereoMIS, moreover, its appearance prior turns vascular texture and specular patches into fragmented depth structure. By comparison, EndoDDC recovers the overall shape of the lumen but leaves pronounced error bands along the folds, and on StereoMIS it over-smooths the scene and preserves errors near the upper image border. EndoStreamDepth reproduces sharp, well-aligned folds on these C3VD frames. On StereoMIS, however, it over-smooths the scene and fails to capture depth transitions near the upper image border, where its error concentrates.

Taken together, these samples demonstrate that TANDEM achieves more balanced performance than all competing methods. Compared with EndoStreamDepth, it yields markedly lower error in flat regions where the memory effectively densifies the anchors, most evidently on StereoMIS, because remembered anchors provide the metric scale that a monocular estimator lacks. Compared with EndoDDC, it substantially improves the edge regions: the residual error of TANDEM on C3VD is confined to short fold segments rather than entire fold contours, and the upper depth transition on StereoMIS is preserved. Compared with Marigold-DC, it avoids the texture-induced artifacts. These gains stem from the design of TANDEM: remembered anchors populate regions lacking current measurements, reliability-weighted integration anchors the global scale to all anchors, and FGRP sharpens the result only near anchor-supported edges. Consequently, TANDEM achieves balanced accuracy across flat regions, depth edges, and both anatomies.

\subsection{Confidence-Aware Temporal Observation Densification}
Table~\ref{tab:density} reports, for four input densities, the number of anchors after priority union, the densification ratio, the median reliability of remembered anchors, and the accuracy. Three key observations emerge. First, densification is gap-driven: with 50 input points, the memory supplies 9.50 (C3VD) and 13.90 (StereoMIS) times as many anchors as the input, whereas with 500 input points, the ratio drops to 3.08 and 4.91, and above the density gate no remembered anchor is injected. Second, reliability is graded rather than binary: the median reliability of remembered anchors rises from 0.328 to 0.630 on C3VD and from 0.260 to 0.507 on StereoMIS as the input becomes denser, because more anchors are re-observed and thus confirmed. with 50 input points, most remembered anchors are of moderate or low reliability; instead of discarding them with a hard threshold, TANDEM attenuates their influence during integration. Third, reliability adapts to the scene without dataset-specific calibration: with identical memory parameters, the medians are lower on StereoMIS. On this dataset, the camera moves more relative to the scene depth and the tissue deforms, so the parallax factor and the re-observation consistency attenuate remembered anchors more strongly and limit contamination from stale anchors. Accuracy improves monotonically with input density on both datasets, down to an RMSE of 0.600\,mm on C3VD and 0.910\,mm on StereoMIS with 50,000 input points. This confirms that the temporal branch does not degrade performance in the dense regime, in which TANDEM operates as a single-frame completion network. Moreover, since every injected anchor carries its reliability, age, and parallax, a clinician or downstream planning module can distinguish whether a depth value relies on a current measurement, a recently confirmed memory entry, or an aging one.

\begin{table}[!t]
\centering
\caption{RMSE (mm) Under Pose Noise (Rotation/Translation Standard Deviation) at Realistic, SLAM-Grade, and Stress Levels. Fixed Omits \eqref{eq:poseaware} and \eqref{eq:poseaware2}; Bare Removes the Reliability Modeling }
\label{tab:pose}
\setlength{\tabcolsep}{3.1pt}
\renewcommand{\arraystretch}{1.05}
\footnotesize
\begin{tabular}{l | c c c | c c c}
\toprule
\multirow{2}{*}{Pose noise} & \multicolumn{3}{c|}{C3VD} & \multicolumn{3}{c}{StereoMIS}\\
 & Pose-aware & Fixed & Bare & Pose-aware & Fixed & Bare\\
\midrule
None & \textbf{1.478} & \textbf{1.478} & 1.657 & \textbf{3.121} & \textbf{3.121} & 3.348\\
0.05$^\circ$/0.15\,mm & \textbf{1.593} & 1.594 & 1.773 & \textbf{3.151} & \textbf{3.151} & 3.387\\
0.1$^\circ$/0.3\,mm & \textbf{1.814} & \textbf{1.814} & 2.021 & 3.216 & \textbf{3.214} & 3.482\\
\midrule
0.2$^\circ$/0.5\,mm & \textbf{2.162} & 2.171 & 2.412 & 3.334 & \textbf{3.331} & 3.646\\
\midrule
0.5$^\circ$/1\,mm & \textbf{3.079} & 3.093 & 3.464 & \textbf{3.676} & \textbf{3.676} & 4.093\\
1$^\circ$/3\,mm & \textbf{5.351} & 5.694 & 6.246 & \textbf{4.811} & 4.942 & 5.600\\
2$^\circ$/5\,mm & \textbf{6.292} & 7.028 & 7.704 & \textbf{5.563} & 5.913 & 6.810\\
\bottomrule
\end{tabular}
\end{table}

\begin{table*}[!t]
\centering
\caption{Ablation Study. Parentheses Give RMSE Changes Relative to the Corresponding Models in (a); RMSE and MAE in mm}
\label{tab:ablation}
\setlength{\tabcolsep}{4.6pt}
\renewcommand{\arraystretch}{1.05}
\footnotesize
\begin{tabular}{l | c c c c | c c c c}
\toprule
\multirow{2}{*}{Configuration} & \multicolumn{4}{c|}{C3VD~\cite{bobrow2023c3vd}} & \multicolumn{4}{c}{StereoMIS~\cite{hayoz2023stereomis}}\\
 & \eh{RMSE $\downarrow$} & \eh{MAE $\downarrow$} & \eh{REL $\downarrow$} & \ah{$\delta$ $\uparrow$} & \eh{RMSE $\downarrow$} & \eh{MAE $\downarrow$} & \eh{REL $\downarrow$} & \ah{$\delta$ $\uparrow$}\\
\midrule
\multicolumn{9}{l}{\emph{(a) Component-wise contribution: components added cumulatively}}\\
Base (single frame) & 2.490 & 1.479 & 0.0599 & 0.9779 & 4.760 & 3.208 & 0.1069 & 0.8736\\
\quad + Memory & 1.543 & 0.874 & 0.0431 & 0.9819 & 4.133 & 2.879 & 0.0935 & 0.9182\\
\quad + Memory + DDI & 1.510 & 0.841 & 0.0379 & 0.9887 & 3.142 & 2.223 & 0.0795 & 0.9491\\
\rowcolor{oursrow}\quad + Memory + DDI + FGRP (TANDEM) & \textbf{1.478} & \textbf{0.823} & \textbf{0.0373} & \textbf{0.9890} & \textbf{3.121} & \textbf{2.205} & \textbf{0.0787} & \textbf{0.9502}\\
\midrule
\multicolumn{9}{l}{\emph{(b) Temporal signal: memory read disabled at inference (same weights as in (a))}}\\
Base + Memory & 4.122 (+167.1\%) & 2.425 & 0.1043 & 0.8891 & 5.971 (+44.5\%) & 4.318 & 0.1362 & 0.7779\\
\quad + DDI & 4.587 (+203.8\%) & 2.669 & 0.1041 & 0.8977 & 4.617 (+46.9\%) & 3.287 & 0.1147 & 0.8795\\
\quad + DDI + FGRP (TANDEM) & 4.586 (+210.3\%) & 2.668 & 0.1040 & 0.8974 & 4.619 (+48.0\%) & 3.279 & 0.1140 & 0.8803\\
\midrule
\multicolumn{9}{l}{\emph{(c) Temporal training: retrained without memory (single-frame input)}}\\
Base + DDI & 2.124 & 1.060 & 0.0440 & 0.9815 & 4.255 & 2.981 & 0.1036 & 0.8954\\
\quad + FGRP & 2.713 & 1.392 & 0.0511 & 0.9791 & 4.302 & 3.007 & 0.1045 & 0.8941\\
\midrule
\multicolumn{9}{l}{\emph{(d) Temporal processing: bare anchor propagation without reliability (same weights as in (a))}}\\
Base + Memory + DDI & 1.675 (+10.9\%) & 0.949 & 0.0425 & 0.9793 & 3.362 (+7.0\%) & 2.360 & 0.0842 & 0.9373\\
\quad + FGRP (TANDEM) & 1.657 (+12.1\%) & 0.929 & 0.0418 & 0.9795 & 3.348 (+7.3\%) & 2.345 & 0.0836 & 0.9379\\
\bottomrule
\end{tabular}
\end{table*}

\subsection{Robustness to Pose Noise}
\label{sec:pose}
TANDEM relies on camera poses to lift and reproject anchors; pose errors therefore constitute a direct source of temporal contamination. Since the poses supplied by the original datasets already contain inherent registration or kinematic errors, the results in Table~\ref{tab:main} inherently reflect these realistic pose‑error conditions. To assess robustness against larger deviations, we additionally inject controlled pose perturbations, covering stress‑test scenarios that exceed the typical error magnitudes observed in the source datasets. We perturbed the poses given to the memory during inference with a rotation about a random axis, whose angle follows a half-normal distribution, and a translation with independent Gaussian components, using the same random seed for all compared models. In addition to the default TANDEM, we evaluated two ablated variants: one using the fixed tolerances defined in Eqs. \eqref{eq:match} and \eqref{eq:rho}, and another implementing bare anchor propagation, in which the memory still injects remembered anchors while reliability modeling is disabled (Section~\ref{sec:ablation}). Since the perturbation is sampled independently per frame, the relative pose between two frames is perturbed by about $\sqrt{2}$ times the reported standard deviation. For the pose-aware default configuration, $(\delta_\theta,\delta_t)$ in \eqref{eq:poseaware} and \eqref{eq:poseaware2} are set to the injected standard deviations, that is, the pose uncertainty is treated as a known specification of the pose source.

\subsubsection{Realistic pose error}
The noise levels are grouped by the pose accuracy reported for endoscopic data. On StereoMIS, whose poses come from the forward kinematics of the da~Vinci Xi, a learned endoscopic pose estimator achieves mean frame-to-frame errors of 0.05$^\circ$ and 0.14\,mm, whereas ORB-SLAM2~\cite{murartal2017orbslam2} reaches 0.20$^\circ$ and 0.33\,mm~\cite{hayoz2023stereomis}. The reference poses of C3VD are themselves registered with mean errors of 0.159$^\circ$ and 0.321\,mm~\cite{bobrow2023c3vd}. We therefore regard per-frame noise up to 0.1$^\circ$ and 0.3\,mm, which is of the same order as these errors, as realistic, 0.2$^\circ$ and 0.5\,mm as the error level of classical SLAM, and larger values as stress tests. As shown in Table~\ref{tab:pose}, the RMSE in the realistic range stays between 1.478 and 1.814\,mm on C3VD and between 3.121 and 3.216\,mm on StereoMIS (at most +22.7\% and +3.0\%). With 0.05$^\circ$ and 0.15\,mm, TANDEM still outperforms every competitor of Table~\ref{tab:main} on C3VD; with 0.1$^\circ$ and 0.3\,mm, only CompletionFormer, which has 15.0 times more parameters, is more accurate, so TANDEM remains on the Pareto front of accuracy and model size. On StereoMIS, TANDEM outperforms all competitors at every level up to 0.5$^\circ$ and 1\,mm.

\subsubsection{Stress tests}
At the SLAM-grade level, the RMSE on C3VD (2.162\,mm) approaches that of the model trained without memory (2.124\,mm), whereas on StereoMIS it remains 21.6\% lower. Beyond this level, remembered anchors become clearly harmful on C3VD from 0.5$^\circ$ and 1\,mm and on StereoMIS from 1$^\circ$ and 3\,mm. While bare anchor propagation yields the highest RMSE consistently, the fixed variant performs comparably to pose‑aware under mild pose noise, with pose‑aware advantages amplifying as perturbations grow severe. The reliability modeling lowers the RMSE of bare propagation by 10.2\%--11.1\% on C3VD and 6.8\%--10.2\% on StereoMIS up to 0.5$^\circ$ and 1\,mm, and by up to 18.3\% at 2$^\circ$ and 5\,mm. The pose-aware tolerances add a gain that grows with the noise, from 0.5\% at 0.5$^\circ$ to 10.5\% at 2$^\circ$ on C3VD and from 2.7\% to 5.9\% between 1$^\circ$ and 2$^\circ$ on StereoMIS, because widened tolerances keep correct matches that fixed tolerances would reject.

\subsection{Ablation Study}
\label{sec:ablation}

Table~\ref{tab:ablation}(a) adds the components cumulatively to the single-frame base network with five-channel input. Here,  Memory denotes the confidence-aware causal temporal memory together with the seven-channel enhanced sparse representation that exposes its evidence; two additional input channels are inherent to this module, as reliability, age, and parallax are provided by the memory. Every component improves all eight metrics on both datasets. The memory provides the largest single gain on C3VD, lowering RMSE, MAE, and REL by 38.0\%, 40.9\%, and 28.0\%, respectively. On StereoMIS, it reduces the RMSE by 13.2\% and the REL by 12.5\% and raises $\delta$ from 0.8736 to 0.9182. Reliability-weighted integration is the decisive component on StereoMIS, where DDI reduces the RMSE by 24.0\% and the MAE by 22.8\%. On C3VD, it lowers the RMSE by 2.1\% and the REL by 12.1\% and raises $\delta$ from 0.9819 to 0.9887. The integration thus ties the wide depth range of in vivo scenes to all anchors, while on the compact colon geometry it mainly reduces the relative error. Finally, FGRP reduces the RMSE by 2.1\% on C3VD and 0.7\% on StereoMIS, with consistent gains in MAE and REL; with only 4.9\,K parameters acting near anchor-supported edges, its effect is local rather than global. Over the complete chain, the RMSE decreases by 40.6\% on C3VD and 34.4\% on StereoMIS relative to the base network.

To test whether the trained models actually depend on remembered anchors, Table~\ref{tab:ablation}(b) disables the memory read at inference for the three temporal configurations of Table~\ref{tab:ablation}(a), keeping their weights and input format unchanged. On C3VD, the RMSE rises from 1.478--1.543\,mm to 4.122--4.587\,mm (+167.1\% to +210.3\%), and on StereoMIS it rises by 44.5\% to 48.0\%. The dependence is strongest for the models with DDI, which use remembered anchors as boundary conditions: without them, the integration is constrained only by the few current measurements, and the gradient field must bridge large unconstrained regions. Because these weights are trained with memory, the numbers quantify how strongly the final models rely on temporal anchors rather than the accuracy attainable without them.

Table~\ref{tab:ablation}(c) retrains the single-frame base network after adding DDI, and after adding DDI and FGRP, without any memory. DDI remains useful without temporal anchors and lowers the RMSE by 14.7\% on C3VD and 10.6\% on StereoMIS. FGRP, in contrast, increases the RMSE by 27.7\% and 1.1\%, respectively, because its gate is driven by the anchor reliability and the input channels for age and parallax, which are constant or absent when no memory exists. Even compared with the best single-frame model, TANDEM lowers the RMSE by 30.4\% (C3VD) and 26.7\% (StereoMIS). Hence, temporal densification is the enabling condition for FGRP and the main source of the accuracy of TANDEM under very sparse input.

AnchorBuffer differs from a plain temporal propagation of points in that it scores every remembered anchor. Table~\ref{tab:ablation}(d) keeps the memory and its anchors but removes this processing at inference: all remembered anchors receive $\tr_t=1$ and zero age and parallax, the reliability-based ordering is replaced by nearest-first ordering, and DDI therefore weights all anchors equally. This bare propagation raises the RMSE of TANDEM by 12.1\% on C3VD and 7.3\% on StereoMIS, and that of the model without FGRP by 10.9\% and 7.0\%. The reliability modeling thus filters contaminated temporal evidence, such as anchors displaced by deformation, parallax, or pose error, that would otherwise be trusted as much as current measurements.

\section{Conclusion}
We presented TANDEM, a lightweight causal framework for sparse depth completion in endoscopic robotic navigation. TANDEM turns temporal redundancy into explicit, reliability-annotated anchors via its AnchorBuffer memory, exposes their provenance to a compact network, and reconstructs depth through reliability-weighted gradient integration and a frequency-gated reference prior. On C3VD and StereoMIS with very sparse input, it outperforms single-frame, diffusion-based, and video-based methods with only 5.58\,M parameters. Its reliability scores make every temporal contribution traceable, and its accuracy degrades only slightly under pose errors of the magnitude reported for endoscopic pose estimation. Future work will couple the memory with estimated camera poses and their uncertainty and validate the framework on clinical sequences.

\section*{Acknowledgment}
The authors would like to express their gratitude to the authors of the C3VD and StereoMIS datasets for making their data publicly available.

\section*{References}
\def\refname{\vadjust{\vspace*{-2.5em}}}
\vspace*{-2.5em}
\bibliographystyle{IEEEtran}
\bibliography{refs}

\end{document}